%% file: main.tex
\documentclass[11pt]{article}

\usepackage[margin=1in]{geometry}
\usepackage{amsmath,amssymb,amsthm}
\usepackage{booktabs}
\usepackage{graphicx}

\usepackage{url}
\usepackage{xcolor}
\usepackage{array}
\usepackage{algorithm}
\usepackage{algpseudocode}
\usepackage{xspace}
\usepackage[hidelinks]{hyperref}
\hypersetup{
  colorlinks=true,
  linkcolor=blue,
  citecolor=blue,
  urlcolor=blue
}

\newtheorem{proposition}{Proposition}
\newtheorem{lemma}{Lemma}
\newtheorem{remark}{Remark}

\newcommand{\FANOS}{\textsc{FANoS-v2}}
\newcommand{\Fast}{\texttt{FANoSV2Fast}}
\newcommand{\FanosRef}{\texttt{FANoSV2}}
\newcommand{\R}{\mathbb{R}}
\DeclareMathOperator{\clip}{clip}

\title{\FANOS: Feedback-Controlled Momentum with Thermostat Damping for Lightweight Neural Optimization}
\author{Nalin Dhiman \\
School of Computing and Electrical Engineering \\
Indian Institute of Technology Mandi \\
d24008@students.iitmandi.ac.in}
\begin{document}
\maketitle

\begin{abstract}
\FANOS{} is a PyTorch optimizer that augments RMS-preconditioned momentum with a scalar feedback controller over update energy. The public reference implementation stores momentum in parameter-update units, applies a non-negative thermostat damping coefficient, supports diagonal, factored, and raw-gradient preconditioning, and exposes diagnostics intended for stability audits. This study gives a complete mathematical specification of the released optimizer, including the exact parameter-unit update, the study-equation physical update mode, bounded log-ratio thermostat control, adaptive preconditioner softening, warmup guardrails, and the experimental \Fast{} profile. We report the v0.2 evidence: five-seed reduced-sample MNIST, Fashion-MNIST, and CIFAR-10 experiments show mean top-1 gains of 0.889, 2.197, and 2.666 percentage points over AdamW for \Fast{}, but with 49.8\%, 61.6\%, and 56.8\% higher wall-clock time. Preliminary scientific, PINN, and EEG smoke tests are mixed and are treated as hypothesis-generating only. The evidence supports \FANOS{} as an alpha-stage research optimizer with a reproducible lightweight-vision signal and an explicit runtime bottleneck. 
\end{abstract}

\section{Introduction}

Adaptive first-order optimizers are often judged against AdamW, a baseline that combines strong empirical behavior, decoupled weight decay, and highly optimized framework implementations \cite{kingma2014adam,loshchilov2019decoupled}. A new optimizer must therefore clear two bars at once. It must show a measurable learning signal, and it must do so without hiding substantial cost or task-specific tuning. \FANOS{} was developed around a specific hypothesis: momentum should not be a fixed inertial constant when the optimizer can directly observe its own update energy. Instead, a lightweight feedback controller can damp momentum when updates become energetically large and relax damping when the update buffer is below a local target. The central design choice in \FANOS{} is to put the controlled state in the same units as the parameter update. Let $u_t$ be the update buffer. The public update is $\theta_{t+1}=\theta_t+u_{t+1}$, where $u_{t+1}$ already contains the learning-rate scaling. This avoids an ambiguity common in early optimizer sketches: whether a velocity state should be added directly to parameters or multiplied by the learning rate first. The repository also retains a physical audit mode in which a descent velocity $v_t$ is stored and the parameter update is $\theta_{t+1}=\theta_t-\eta_t v_{t+1}$. The public default is the parameter-unit mode because its learning-rate semantics match ordinary deep-learning optimizers. This study provides a quality mathematical and experimental record for the v0.2 benchmarked state. The strongest current empirical result is repeated-seed lightweight vision: on reduced-sample MNIST, Fashion-MNIST, and CIFAR-10 runs, \Fast{} improves mean top-1 accuracy over AdamW in the project evidence. The weakness is equally clear: \Fast{} remains approximately 50-62\% slower than AdamW on those same experiments. The scientific and EEG smoke tests are useful for debugging and prioritization, but not yet strong enough for broad claims.

\paragraph{Contributions.}
This study contributes: (i) a complete mathematical specification of \FANOS{} and \Fast{}; (ii) stability-oriented lemmas explaining what the controller does and does not guarantee; (iii) a critical empirical report using the benchmark summaries.

\section{Claim Boundary}

The present v0.2 release supports the following claim:
\begin{quote}
\FANOS{} is a feedback-controlled optimization framework that shows repeated-seed accuracy improvements on lightweight vision benchmarks, together with a technically plausible route toward scientific and physics-informed learning objectives.
\end{quote}

We deliberately distinguish this supported claim from stronger conclusions that are not yet established by the current evidence. In particular, the v0.2 experiments do not demonstrate that \FANOS{} is a universal replacement for AdamW, that it is currently competitive in wall-clock efficiency across modern training regimes, or that it has achieved broad empirical superiority on physics-informed neural network benchmarks. Establishing those claims will require larger-scale studies, broader task coverage, more systematic runtime profiling, and independent replication. Accordingly, we position \FANOS{} v0.2 as an alpha-stage research optimizer. The reference implementation, \FanosRef{}, is intended to provide a mathematically transparent and auditable version of the update rule. The speed-oriented implementation, \Fast{}, represents the current v0.2 practical profile: its accuracy signal has been revalidated in the archived lightweight-vision runs, while its computational overhead remains the primary target for future engineering and algorithmic refinement.

\section{Related Work}

\paragraph{Momentum and adaptive gradients.}
Classical heavy-ball momentum accelerates optimization by accumulating a velocity-like state \cite{polyak1964some}. AdaGrad adapts coordinates through accumulated gradient magnitudes \cite{duchi2011adaptive}, RMSProp uses an exponential moving average of squared gradients \cite{tieleman2012lecture}, and Adam combines first and second moments with bias correction \cite{kingma2014adam}. AdamW decouples weight decay from the adaptive gradient step and remains a strong default for neural-network training \cite{loshchilov2019decoupled}. \FANOS{} is closest to RMS-preconditioned momentum, but replaces fixed momentum behavior with a feedback-controlled damping coefficient.

\paragraph{Memory-aware preconditioning.}
Factored second-moment methods such as Adafactor and tensor preconditioners such as Shampoo reduce or reshape optimizer state for large tensors \cite{shazeer2018adafactor,gupta2018shampoo}. \FANOS{} currently supports a simple factored RMS state for matrix-like tensors, but it still stores a full update buffer. Therefore, its present factored mode is a partial memory reduction, not a complete low-memory optimizer.

\paragraph{Thermostat inspiration.}
The term thermostat is borrowed from controlled-temperature molecular dynamics, especially Nose-Hoover style schemes \cite{nose1984unified,hoover1985canonical}. \FANOS{} is not a Hamiltonian sampler and does not claim canonical sampling. It uses a first-order stochastic optimizer update with a clipped log-ratio controller that reacts to update energy.

\paragraph{Scientific neural optimization.}
Physics-informed neural networks can be sensitive to loss scaling, boundary terms, residual sampling, and optimizer handoff strategies \cite{raissi2019physics}. The \FANOS{} PINN evidence in this package is a single Poisson-1D preset result plus a mixed smoke-suite result. It motivates further testing but does not justify a broad PINN claim.

\section{Notation and Objective}

Let $f(\theta)=\mathbb{E}_{\xi}[\ell(\theta;\xi)]$ be a differentiable objective over parameters $\theta\in\R^d$. At step $t$, the optimizer receives a stochastic gradient $g_t$. The learning rate is $\eta_t>0$. The second-moment decay is $\beta_2\in[0,1)$. The RMS stabilizer is $\epsilon>0$. The base momentum coefficient after warmup is $m\in[0,1)$. The scalar thermostat state for a parameter group is $\zeta_t$, clipped to $[\zeta_{\min},\zeta_{\max}]$; the public default uses $\zeta_{\min}=0$, so the controller only removes momentum energy rather than injecting negative friction.

The optimizer supports multiple parameter groups. For clarity, the mathematical specification below is written for one group. The implementation applies the same equations independently to each group, with group-level scalar diagnostics and per-parameter tensor state.

\section{FANoS-v2 Algorithm}

\subsection{Diagonal RMS-preconditioned parameter-unit update}

The default preconditioner is a diagonal exponential moving average of squared gradients:
\begin{align}
  s_t &= \beta_2 s_{t-1} + (1-\beta_2) g_t^2, \\
  \widehat{s}_t &= \frac{s_t}{1-\beta_2^t}, \\
  d_t &= \left(\sqrt{\widehat{s}_t}+\epsilon\right)^{\alpha_t}, \\
  \widetilde{g}_t &= g_t \oslash d_t .
\end{align}
Here $\alpha_t\geq 0$ is the preconditioner power. The default $\alpha_t=1$ gives full RMS preconditioning. The value $\alpha_t=0.5$ gives softer RMS preconditioning, and $\alpha_t=0$ reduces to raw-gradient feedback momentum. Bias correction is enabled by default because otherwise the first steps can be too large when $\beta_2$ is close to one.

The momentum coefficient is damped by the thermostat:
\begin{align}
  \rho_t &= \clip\left(m_t \exp(-\eta_t\zeta_t),\,0,\,0.999\right),
\end{align}
where $m_t$ is the warmup-adjusted base momentum. The public update buffer $u_t$ is in parameter units:
\begin{align}
  u_{t+1} &= \rho_t u_t - \eta_t\widetilde{g}_t, \\
  \theta_{t+1} &= \theta_t + u_{t+1}. \label{eq:param-update}
\end{align}
The sign convention is therefore ordinary descent: if $u_t=0$ and $\eta_t>0$, then the new step is negative in the preconditioned gradient direction.

\subsection{Physical audit mode}

For study-equation audits, the repository supports \texttt{update\_mode="physical"}. In that mode, the state is a descent velocity $v_t$:
\begin{align}
  v_{t+1} &= \rho_t v_t + \widetilde{g}_t, \\
  \theta_{t+1} &= \theta_t - \eta_t v_{t+1}. \label{eq:physical-update}
\end{align}
When $u_t=-\eta_t v_t$ and $\eta_t$ is constant, the two modes are closely related. The parameter-unit mode is preferred for public training because the learning rate enters the update exactly once.

\subsection{Thermostat control law}

The controller compares observed update energy with a target based on the proposed preconditioned step. Let
\begin{align}
  K_t &= \frac{1}{d}\|\Delta\theta_t\|_2^2, \\
  K_t^\star &= c\frac{1}{d}\|\eta_t\widetilde{g}_t\|_2^2 + \epsilon,
\end{align}
where $c>0$ is \texttt{target\_scale}. In parameter mode, $\Delta\theta_t=u_{t+1}$. In physical mode, $\Delta\theta_t=-\eta_t v_{t+1}$. The implementation forms exponential moving averages
\begin{align}
  \bar K_t &= b \bar K_{t-1} + (1-b)K_t, \\
  \bar K_t^\star &= b \bar K_{t-1}^\star + (1-b)K_t^\star,
\end{align}
with initialization from the first observed control step. The log-ratio error is
\begin{align}
  e_t = \clip\left(\log\frac{\bar K_t+\epsilon}{\bar K_t^\star+\epsilon},\,-e_{\max},\,e_{\max}\right). \label{eq:logerr}
\end{align}
After thermostat warmup, the scalar damping is updated as
\begin{align}
  \zeta_{t+1}=\clip\left((1-\lambda)\zeta_t+\gamma e_t,\,\zeta_{\min},\,\zeta_{\max}\right), \label{eq:zeta-update}
\end{align}
where $\gamma$ is \texttt{thermostat\_lr} and $\lambda$ is \texttt{thermostat\_decay}. The default lower bound $\zeta_{\min}=0$ is deliberate. If the update energy is too high, $e_t>0$ and the controller increases damping. If the update energy is too low, $e_t<0$ and the controller can reduce damping, but it does not become negative under the default bounds.

\subsection{Warmup, adaptive learning rate, and adaptive preconditioner power}

The implementation includes guardrails that are useful in short neural-network runs. With warmup length $T_w$, the scalar
\begin{align}
  w_t = \clip(t/T_w,0,1)
\end{align}
interpolates base momentum from \texttt{warmup\_start\_momentum} to the requested momentum. A separate warmup can delay thermostat updates, keeping $\zeta_t$ fixed during early training.

The optional adaptive learning-rate scalar uses a gradient-norm EMA:
\begin{align}
  r_t &= \log\frac{\|g_t\|_2+\delta}{\overline{\|g\|}_{t}+\delta}, \\
  \eta_t &= \clip\left(\eta\exp(-\kappa r_t),\,\eta_{\min},\,\eta_{\max}\right),
\end{align}
when \texttt{adaptive\_lr=True}. This mechanism is disabled in \Fast{} because the v0.2 speed profile removes gradient-norm adaptive learning rate and clipping.

The optional adaptive preconditioner power softens the RMS exponent when recent thermostat error indicates instability. In the implementation,
\begin{align}
  \alpha_t = \clip\left(\alpha_{\rm sched}(t) - q\max(0, |e_{t-1}|-1),\,\alpha_{\min},\,\alpha_{\max}\right).
\end{align}
The \texttt{auto} preset uses bounds $[0.5,1.0]$ and target $1.0$. This keeps ordinary startup close to RMS/Adam-style preconditioning while allowing the controller to soften toward $0.5$ under large energy error.

\subsection{Factored and raw-gradient preconditioners}

The diagonal mode stores one full second-moment tensor and one full update buffer. The raw-gradient mode stores only the update buffer and sets $\widetilde g_t=g_t$. The factored mode is used for matrix-like tensors. For a matrix gradient $G_t\in\R^{r\times c}$, it maintains row and column second-moment factors:
\begin{align}
  a_t &= \beta_2 a_{t-1} + (1-\beta_2)\operatorname{mean}_{j}(G_t^2[:,j]), \\
  b_t &= \beta_2 b_{t-1} + (1-\beta_2)\operatorname{mean}_{i}(G_t^2[i,:]).
\end{align}
After bias correction, the factored denominator is computed as
\begin{align}
  D_t^2 = \frac{a_t b_t^\top}{\operatorname{mean}(a_t)},
\end{align}
then $D_t=(\sqrt{D_t^2}+\epsilon)^{\alpha_t}$. For tensors whose rank is below the configured threshold, factored mode falls back to the diagonal state. Because the update buffer remains full-size, factored mode reduces second-moment memory but not momentum/update memory.

\subsection{Pseudocode}
\begin{algorithm}[H]
\small
\caption{Reference \FanosRef{} parameter-unit update for one parameter group}
\label{alg:fanos-ref-step}
\begin{algorithmic}
\Require Parameters $\theta$, gradients $g_t$, learning rate $\eta$, second-moment coefficient $\beta_2$, numerical constant $\epsilon$, base momentum $m$, thermostat state $\zeta_t$, target scale $c$, optimizer state $(u_t,s_t)$
\Ensure Updated $\theta_{t+1}$, velocity $u_{t+1}$, second-moment state $s_{t+1}$, thermostat state $\zeta_{t+1}$

\State $t \gets t+1$

\If{\textsc{SanitizeGradients} is enabled}
    \State $g_t \gets \mathrm{where}(\mathrm{isfinite}(g_t), g_t, 0)$
\EndIf

\If{gradient clipping is enabled}
    \State $G_t \gets \|g_t\|_2$
    \State $\alpha_t \gets \min(1,\tau/(G_t+\epsilon))$
    \State $g_t \gets \alpha_t g_t$
\EndIf

\If{coupled weight decay is enabled}
    \State $g_t \gets g_t + \lambda \theta_t$
\ElsIf{decoupled weight decay is enabled}
    \State $\theta_t \gets (1-\eta\lambda)\theta_t$
\EndIf

\If{diagonal preconditioning is enabled}
    \State $s_{t+1} \gets \beta_2 s_t + (1-\beta_2) g_t^2$
    \State $\widetilde g_t \gets g_t / (\sqrt{s_{t+1}}+\epsilon)$
\ElsIf{factored preconditioning is enabled}
    \State Update factored second-moment statistics from $g_t^2$
    \State Construct factored preconditioner $P_t$
    \State $\widetilde g_t \gets P_t g_t$
\Else
    \State $\widetilde g_t \gets g_t$
\EndIf

\State $\eta_t \gets \textsc{WarmupLR}(\eta,t)$
\State $m_t \gets \textsc{WarmupMomentum}(m,t)$
\State $\rho_t \gets \mathrm{clip}\!\left(m_t \exp(-\eta_t \zeta_t),0,0.999\right)$

\State $u_{t+1} \gets \rho_t u_t - \eta_t \widetilde g_t$
\State $\theta_{t+1} \gets \theta_t + u_{t+1}$

\If{$t$ is a thermostat-control step}
    \State $K_t \gets \|u_{t+1}\|_2^2$
    \State $K_t^\star \gets c\,\eta_t^2 \|\widetilde g_t\|_2^2$
    \State $\overline K_t \gets \textsc{EMA}(K_t)$
    \State $\overline K_t^\star \gets \textsc{EMA}(K_t^\star)$
    \State $e_t \gets \log(\overline K_t+\epsilon)-\log(\overline K_t^\star+\epsilon)$
    \If{thermostat warmup is complete}
        \State $\zeta_{t+1} \gets \textsc{ThermostatUpdate}(\zeta_t,e_t)$
    \Else
        \State $\zeta_{t+1} \gets \zeta_t$
    \EndIf
\Else
    \State $\zeta_{t+1} \gets \zeta_t$
\EndIf

\If{diagnostics are enabled}
    \State Record $(t,\zeta_{t+1},\rho_t,K_t,K_t^\star,e_t,G_t,\alpha_t,\eta_t)$
\EndIf

\State \Return $\theta_{t+1},u_{t+1},s_{t+1},\zeta_{t+1}$
\end{algorithmic}
\end{algorithm}

\begin{algorithm}[H]
\caption{Speed-oriented \Fast{} v0.2 profile}
\label{alg:fanos-fast-profile}
\begin{algorithmic}
\Require Reference \FanosRef{} configuration $\mathcal C_{\mathrm{ref}}$
\Ensure Speed-oriented configuration $\mathcal C_{\mathrm{fast}}$

\State $\mathcal C_{\mathrm{fast}} \gets \mathcal C_{\mathrm{ref}}$

\If{\texttt{preset} is unspecified}
    \State $\mathcal C_{\mathrm{fast}}.\texttt{preset} \gets \texttt{"auto"}$
\EndIf

\State $\mathcal C_{\mathrm{fast}}.\texttt{adaptive\_lr} \gets \texttt{False}$
\State $\mathcal C_{\mathrm{fast}}.\texttt{grad\_clip} \gets \texttt{None}$
\State $\mathcal C_{\mathrm{fast}}.\texttt{thermostat\_interval} \gets 4$
\State $\mathcal C_{\mathrm{fast}}.\texttt{sanitize\_gradients} \gets \texttt{True}$
\State $\mathcal C_{\mathrm{fast}}.\texttt{record\_diagnostics} \gets \texttt{False}$

\For{each training step $t$}
    \State Apply Algorithm~\ref{alg:fanos-ref-step} using $\mathcal C_{\mathrm{fast}}$
\EndFor

\State \Return $\mathcal C_{\mathrm{fast}}$
\end{algorithmic}
\end{algorithm}

\newpage
\section{Theoretical Properties and Limits}

This section gives stability-oriented facts that follow from the update equations. The controller includes clipping, warmup, finite-precision state, optional adaptive learning-rate scalars, and optional adaptive preconditioner power; those choices are useful engineering guardrails but make a clean theorem less meaningful than a boundedness and sign analysis.

\begin{lemma}[Bounded damping and momentum factor]
Assume $m_t\in[0,m_{\max}]$ with $m_{\max}<1$, $\eta_t>0$, and $\zeta_t\geq 0$. Then the effective momentum factor satisfies $0\leq \rho_t\leq m_{\max}<1$ before the implementation's additional cap at $0.999$.
\end{lemma}
\begin{proof}
Because $\zeta_t\geq0$ and $\eta_t>0$, $\exp(-\eta_t\zeta_t)\in(0,1]$. Hence $m_t\exp(-\eta_t\zeta_t)\leq m_t\leq m_{\max}$. Clipping to a non-negative interval cannot increase the upper bound beyond the cap.
\end{proof}

\begin{lemma}[One-step descent direction without momentum]
In parameter-unit mode, suppose $u_t=0$, $\eta_t>0$, and each preconditioner denominator coordinate is positive. Then the step direction $\Delta\theta_t=-\eta_t D_t^{-1}g_t$ satisfies $g_t^\top\Delta\theta_t\leq0$, with equality only for zero gradient coordinates under positive denominators.
\end{lemma}
\begin{proof}
The inner product is $g_t^\top(-\eta_t D_t^{-1}g_t)=-\eta_t\sum_i g_{t,i}^2/D_{t,i}$, which is non-positive because $\eta_t>0$ and $D_{t,i}>0$.
\end{proof}

\begin{proposition}[Bounded update buffer under bounded preconditioned gradients]
Assume parameter-unit mode, $\rho_t\leq\bar\rho<1$, $\eta_t\leq\eta_{\max}$, and $\|\widetilde g_t\|_2\leq B$ for all $t$. Then
\begin{align}
  \|u_t\|_2 \leq \bar\rho^t\|u_0\|_2 + \frac{\eta_{\max}B}{1-\bar\rho}.
\end{align}
\end{proposition}
\begin{proof}
From $u_{t+1}=\rho_tu_t-\eta_t\widetilde g_t$ and the triangle inequality, $\|u_{t+1}\|_2\leq\bar\rho\|u_t\|_2+\eta_{\max}B$. Unrolling the scalar recurrence gives the result.
\end{proof}

\begin{proposition}[Controller sign]
Assume a control step after thermostat warmup and ignore clipping saturation. If $\bar K_t>\bar K_t^\star$, then $e_t>0$ and the controller increases $\zeta$ when $\gamma>0$ and $\lambda=0$. This lowers the next effective momentum factor for fixed $m$ and $\eta$. If $\bar K_t<\bar K_t^\star$, then $e_t<0$ and the controller reduces $\zeta$ until it reaches the lower bound.
\end{proposition}
\begin{proof}
The sign of $e_t$ is the sign of the log-ratio in Eq.~\eqref{eq:logerr}. With $\lambda=0$, Eq.~\eqref{eq:zeta-update} adds $\gamma e_t$ to $\zeta_t$ before clipping. The derivative of $m\exp(-\eta\zeta)$ with respect to $\zeta$ is $-\eta m\exp(-\eta\zeta)<0$, so increasing $\zeta$ lowers $\rho$.
\end{proof}

\begin{remark}
The lemmas describe boundedness and feedback direction. They do not imply that $f(\theta_t)$ decreases every step under stochastic gradients, nonconvex losses, changing preconditioners, or finite-batch noise. They also do not prove that the thermostat target is optimal. The empirical section therefore remains essential.
\end{remark}

\section{Implementation Details}

The implementation is a PyTorch optimizer with one scalar thermostat state per parameter group and tensor state per parameter. The main public class is \FanosRef{}. The class \Fast{} subclasses \FanosRef{} and changes only runtime defaults. Tests in the source snapshot check finite-loss smoke behavior, quadratic convergence from a large initial point, non-negative bounded $\zeta$, clipping before second-moment updates, first-step bias correction, raw-gradient mode, auto-preset override behavior, PINN preset defaults, thermostat interval behavior, diagnostic intervals, state-dictionary round trips, factored preconditioner state, physical update sign, adaptive learning-rate bounds, and memory-helper round trips. The optimizer also exports experimental helper functions for low-rank approximation, packed signed 4-bit quantization, top-$k$ sparsification, dynamic variance clipping, distributed gradient averaging, and quantization residual analysis. These helpers are intentionally outside the optimizer step. They are useful for memory and communication experiments but should not be presented as part of the core convergence claim.

\paragraph{State memory.}
For $n$ trainable scalar parameters, diagonal \FANOS{} stores a full update buffer and a full second-moment buffer, giving approximately $2n$ optimizer-state scalars, similar in count to AdamW's two moment tensors. Raw-gradient mode stores only the update buffer. Factored mode stores the full update buffer plus row/column second-moment factors for eligible matrix-like tensors. Optional lower-precision state can reduce byte usage.

\paragraph{Runtime bottleneck.}
The current implementation is Python-loop heavy. AdamW in PyTorch has optimized dense paths, while \FANOS{} performs extra scalar logic and per-parameter tensor operations. The v0.2 fast profile removes diagnostic object creation and gradient-norm adaptive learning rate, but the final vision runs still show substantial overhead. The next engineering milestone should be a dense diagonal fast path using batched or foreach-style tensor operations.

\section{Experimental Protocol}

All quantitative results reported in this manuscript are drawn from the archived CSV summaries included with the release package. The purpose of the experimental section is to document the v0.2 optimization signal under controlled, reproducible settings. We therefore frame the results as an optimizer-behavior study.

\subsection{Vision tasks}

The fast-profile validation uses MNIST \cite{lecun1998gradient}, Fashion-MNIST \cite{xiao2017fashion}, and CIFAR-10 \cite{krizhevsky2009learning}. Each task is evaluated over five random seeds using 10{,}000 training samples, 2{,}000 test samples, five training epochs, and the repository's small-CNN benchmark protocol. AdamW and \Fast{} are compared under matched data splits, architectures, seeds, batch sizes, and benchmark-script defaults. These lightweight experiments are designed to test whether the optimizer produces a repeatable accuracy signal under controlled conditions. They provide a reproducible first-pass validation that motivates broader evaluation.

\subsection{Ablations}

The release includes scalar-synchronization ablations that vary the gradient-norm refresh interval. During development, we identified an interaction in which the \texttt{auto} preset could override an explicit request to disable adaptive learning-rate control. This interaction was corrected, regression tests were added, and the final \Fast{} default was revalidated after the fix. This correction is important for interpretability: the final fast-profile results correspond to the intended configuration and do not rely on hidden retention of gradient-norm adaptive learning-rate behavior.

\subsection{Scientific, PINN, and EEG checks}

The scientific smoke suite contains Rosenbrock-100, noisy small regression, exponential ODE fitting, sequence memory, an ill-conditioned quadratic task, and a Poisson-1D PINN. The package also includes a preliminary EEGBCI cross-subject smoke result benchmark. EEGBCI-related tooling follows the EEG Motor Movement/Imagery dataset and standard BCI evaluation conventions \cite{goldberger2000physiobank,schalk2004bci2000}.

These experiments serve as diagnostic coverage beyond lightweight vision. They test whether the optimizer behaves sensibly across heterogeneous loss geometries, but they are not presented as definitive evidence of broad superiority in scientific machine learning, PINNs, or EEG modeling. Their role is to identify promising regimes and failure modes for the next evaluation cycle.

\section{Results}

\subsection{Final fast vision validation}

Table~\ref{tab:vision-fast} reports the final five-seed reduced-sample vision validation. \Fast{} improves mean top-1 on all three tasks, with the largest absolute improvement on CIFAR-10. It is also slower on all three tasks.

\begin{table}[t]
\centering
\caption{Final five-seed reduced-sample vision validation. Top-1 values are percentages. Runtime is wall-clock seconds reported by the benchmark script.}
\label{tab:vision-fast}
\resizebox{\textwidth}{!}{\input{tables/vision_fast_results.tex}}
\end{table}

\begin{figure}[t]
\centering
\includegraphics[width=0.82\textwidth]{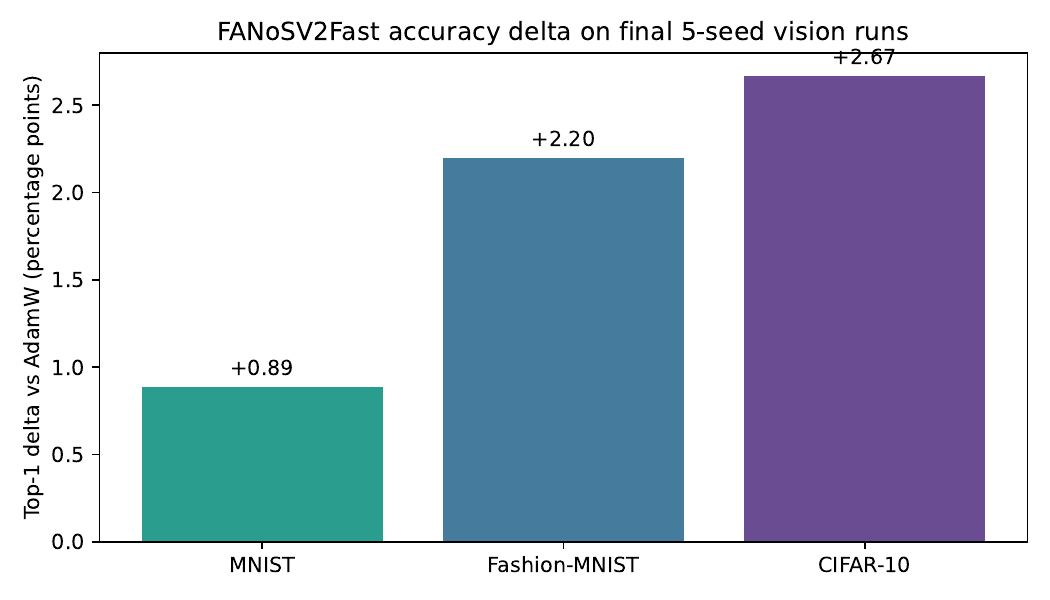}
\caption{Top-1 accuracy delta of \Fast{} relative to AdamW on final five-seed vision runs.}
\label{fig:vision-accuracy}
\end{figure}

\begin{figure}[t]
\centering
\includegraphics[width=0.82\textwidth]{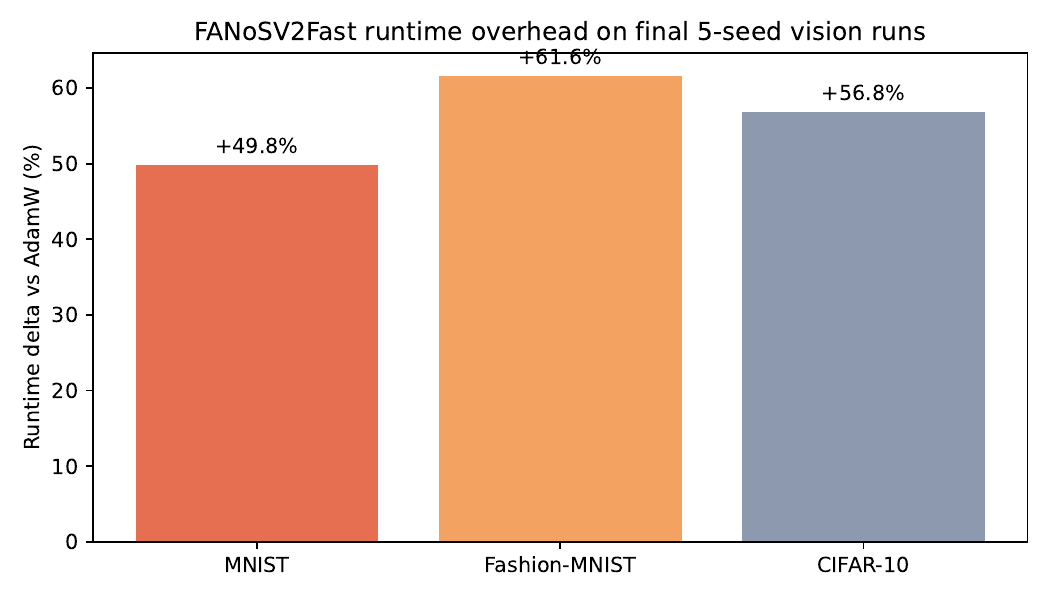}
\caption{Runtime overhead of \Fast{} relative to AdamW on final five-seed vision runs. The overhead remains the central blocker for practical adoption.}
\label{fig:vision-runtime}
\end{figure}

The accuracy signal is encouraging because it survives the fast-profile simplification. However, the cost is too large to ignore. A practical optimizer release must either narrow this runtime gap or provide a use case where the accuracy gain is worth the cost.

\subsection{Gradient-norm synchronization ablation}

Table~\ref{tab:gradnorm} summarizes the gradient-norm refresh interval ablation. The main lesson is that interval tuning alone was not a clean solution. MNIST showed a speed improvement at larger intervals but with reduced or more variable accuracy; CIFAR-10 did not justify promoting interval tuning as the main v0.2 result.

\begin{table}[t]
\centering
\caption{Gradient-norm refresh interval ablation. Values are five-seed summaries against the same AdamW baseline in each folder.}
\label{tab:gradnorm}
\resizebox{\textwidth}{!}{\input{tables/gradnorm_ablation.tex}}
\end{table}

\begin{figure}[t]
\centering
\includegraphics[width=0.82\textwidth]{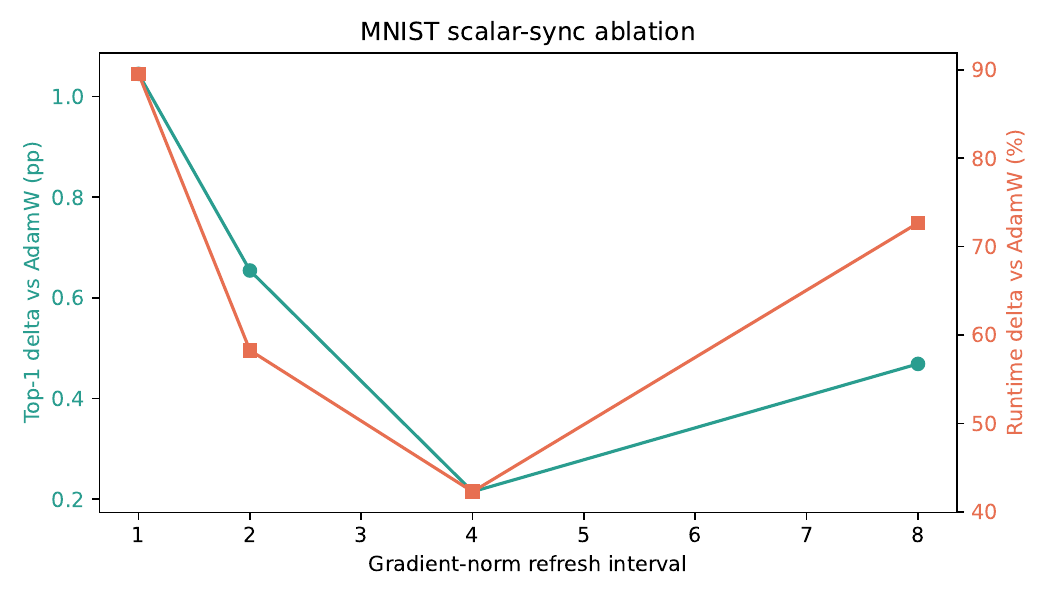}
\caption{MNIST interval sweep. Higher intervals reduce some scalar synchronization but make training behavior less predictable.}
\label{fig:gradnorm}
\end{figure}

\subsection{Preliminary PINN result}

Table~\ref{tab:pinn} reports the cleaner Poisson-1D PINN preset result. The task-aware \FANOS{} configuration has a lower residual metric than AdamW in this one setting, but it is over three times slower. The broader smoke-suite table below contains a different, worse Poisson entry for generic \FANOS{}; that conflict is important. The task-aware preset can help in a narrow setup, but generic scientific advantage is not established.

\begin{table}[t]
\centering
\caption{Preliminary Poisson-1D PINN preset result. Lower residual metric is better. This is a single task and does not establish broad PINN superiority.}
\label{tab:pinn}
\resizebox{0.82\textwidth}{!}{\input{tables/pinn_preliminary.tex}}
\end{table}

\subsection{Preliminary EEG result}

Table~\ref{tab:eeg} documents the EEG smoke benchmark. \FANOS{} auto has a small mean top-1 edge over AdamW, but the variance is high and the time overhead is large. This should be treated as a fixed-benchmark sanity check.

\begin{table}[t]
\centering
\caption{Preliminary EEGBCI five-seed result. This table is included to document the fixed benchmark, not to claim EEG superiority.}
\label{tab:eeg}
\resizebox{0.82\textwidth}{!}{\input{tables/eeg_preliminary.tex}}
\end{table}

\subsection{Scientific smoke suite}

The small scientific suite is mixed. \FANOS{} is competitive or useful on some tasks, but not uniformly better than AdamW or RMSProp. The ill-conditioned quadratic entry failed in the fixed full-study summary and remains an explicit v0.3 target. On Rosenbrock-100, RMSProp outperforms both AdamW and \FANOS{} in the archived metric. On sequence memory, \FANOS{} and AdamW both reach perfect accuracy, but \FANOS{} is slower. These results argue for tighter task-specific hypotheses, not broad claims.

\begin{table}[t]
\centering
\caption{Selected scientific smoke results from the fixed full-study run. Lower metric is better except sequence-memory accuracy where higher is better. This is diagnostic evidence, not a broad scientific-ML benchmark.}
\label{tab:scientific}
\resizebox{\textwidth}{!}{\input{tables/scientific_preliminary.tex}}
\end{table}

\section{Discussion}

The v0.2 result is meaningful because the final fast validation isolates a clearer signal than earlier drafts. The invalid no-gradnorm run was corrected, the fast profile was made explicit, and the final validation still showed accuracy gains on three lightweight vision datasets. This suggests that the feedback-controlled momentum structure deserves further study.

The same evidence also shows why the release must be careful. The optimizer is not currently faster than AdamW, and the cost is not a small constant. The overhead is consistent with the implementation structure: more scalar control, more Python iteration, and fewer fused dense operations. Therefore, the next best research step is not to add many more benchmark claims. It is to engineer a dense diagonal fast path, rerun the same three vision tasks, and check whether the accuracy signal survives at lower cost.

The scientific story needs even more restraint. The Poisson-1D preset result is promising, but the generic smoke-suite Poisson entry is poor. PINN behavior depends on collocation sampling, residual and boundary weighting, architecture, activation, normalization, and optimizer handoff. A serious follow-up should include multiple PDEs, AdamW+LBFGS baselines, validation-grid error, residual terms, runtime, failure rates, and enough seeds to estimate variability.
\section{Threats to Validity}

\paragraph{Benchmark scale.}
The final vision experiments use reduced-sample small-CNN benchmarks. This setting is appropriate for controlled optimizer-behavior validation, but it does not establish transfer to substantially larger regimes such as ResNets, Vision Transformers, language models, diffusion models, large-batch training, or distributed optimization.

\paragraph{Runtime measurement.}
Runtime values are taken from the archived benchmark-script summaries. These measurements are useful for diagnosing the current v0.2 implementation profile, especially relative overhead under matched scripts. A complete efficiency study should additionally include profiler traces, explicit synchronization accounting, kernel-level attribution, memory-bandwidth analysis, and hardware-specific reporting.

\paragraph{Hyperparameter coverage.}
The comparisons use project defaults and documented practical presets. They are intended to evaluate the released configurations rather than to perform exhaustive optimizer tuning. A stronger comparative study would jointly tune AdamW, SGD, RMSProp, \FANOS{}, and related adaptive methods across learning rates, momentum settings, weight decay, scheduler choices, and task-specific training budgets.

\paragraph{Statistical power.}
The main lightweight-vision experiments use five seeds per task. This is sufficient to expose large instabilities and repeated-seed directional effects, but it is not a substitute for a high-power statistical study of small differences. Future evaluations should report confidence intervals, paired significance tests where applicable, complete per-seed traces, and sensitivity to data subsampling.

\paragraph{Implementation maturity.}
The v0.2 implementation includes tests, diagnostics, reference update logic, and a speed-oriented profile. However, it remains an alpha-stage research implementation. In particular, it does not yet include a fully optimized \texttt{foreach} or fused-kernel path, and it has not been extensively stress-tested under mixed precision, distributed data parallelism, sparse gradients, gradient accumulation, or very large model training.

\section{Software Availability and Release Positioning}

\FANOS{} is released as an alpha-stage PyTorch research optimizer for feedback-controlled momentum experiments. The package is publicly available on PyPI and can be installed with:
\begin{center}
\texttt{pip install fanos}
\end{center}
The source code, examples, benchmark scripts, and release materials are maintained at:
\begin{center}
\url{https://github.com/nalin-dhiman/fanos}
\end{center}
The PyPI package page is:
\begin{center}
\url{https://pypi.org/project/fanos/}
\end{center}

The release exposes two intended usage profiles. The first is \FanosRef{}, the reference implementation, which is designed for mathematical transparency, auditability, diagnostics, and ablation studies. The second is \Fast{}, the speed-oriented v0.2 profile, which reduces diagnostic and scalar-synchronization overhead while remaining within the same update family.

The current evidence supports the use of \FANOS{} for optimizer research, controlled benchmarking, and exploratory studies of feedback-regulated momentum. In the archived v0.2 experiments, \Fast{} shows repeated-seed lightweight-vision accuracy gains over AdamW under the repository benchmark protocol. At the same time, the present implementation carries nontrivial runtime overhead, and the available evidence does not yet justify treating \FANOS{} as a default replacement for AdamW in production training pipelines. Accordingly, the release is positioned as a research optimizer rather than a production-optimized training backend. The repository includes the archived CSV summaries, generated tables, figure scripts, benchmark entry points, and diagnostic outputs used to support the results reported in this manuscript. Scientific-ML, PINN, and EEG results are included as preliminary diagnostic checks and should be interpreted as motivation for broader evaluation rather than as final task-level conclusions.

The immediate development priority for the next release is performance engineering. In particular, v0.3 should focus on reducing scalar synchronization, improving \texttt{foreach} or fused-kernel support, profiling mixed-precision behavior, and validating the optimizer under larger models and longer training budgets before making broader benchmark claims.

\section{Conclusion}

\FANOS{} reaches a credible v0.2 milestone: a mathematically auditable optimizer, a bounded feedback controller, an explicit fast profile, and repeated-seed lightweight-vision gains over AdamW in the archived evidence. Runtime overhead remains the practical blocker, scientific and EEG evidence is preliminary, and the current theory establishes stability properties rather than broad convergence superiority. The strongest next step is implementation engineering followed by the same controlled validation.
\appendix

\section{Default Parameters and Presets}

The reference constructor exposes learning rate, $\beta_2$, $\epsilon$, momentum, thermostat learning rate, thermostat decay, EMA coefficient, target scale, $\zeta$ bounds, maximum log error, gradient clipping, weight decay, preconditioner choice, preconditioner power, state dtype, bias correction, update mode, adaptive learning-rate controls, warmup controls, thermostat interval, gradient-norm interval, gradient sanitization, and diagnostics controls. The \texttt{auto} preset sets guarded startup defaults: momentum remains 0.85 unless overridden, target scale remains 0.10 unless overridden, thermostat learning rate remains 0.003 unless overridden, adaptive learning rate defaults to true, adaptive preconditioner power is enabled with bounds $[0.5,1.0]$, momentum warmup defaults to 200 steps, and thermostat warmup defaults to 100 steps. The \texttt{pinn} preset uses softer task-specific defaults: learning rate $5\times10^{-4}$, momentum 0.75, target scale 0.05, thermostat learning rate 0.001, adaptive learning rate enabled, preconditioner power 0.5, 200-step momentum warmup, and 200-step thermostat warmup unless explicitly overridden. The \Fast{} profile starts from \texttt{auto}, disables adaptive learning rate and gradient clipping, uses a thermostat interval of four, keeps gradient sanitization enabled, and disables diagnostics by default.

\section{Additional Implementation Notes}

\paragraph{Clipping order.} When gradient clipping is enabled, clipping occurs before the second-moment update. This prevents a large outlier gradient from permanently inflating the RMS state before clipping is applied.

\paragraph{Bias correction.} With $\beta_2$ close to one, uncorrected $s_t$ is initially small. Bias correction makes the first preconditioned step comparable to AdamW-style behavior and is tested explicitly.

\paragraph{Sparse gradients.} The optimizer currently rejects sparse gradients. This is a deliberate safety choice for the alpha implementation.

\paragraph{Diagnostics.} Diagnostics are group-level snapshots, not per-tensor traces. They are intended to audit controller behavior without storing large histories.

\bibliographystyle{plain}
\bibliography{references}

\end{document}

%% file: tables/vision_fast_results.tex
\begin{tabular}{lrrrrrr}
\toprule
Dataset & AdamW top-1 & FANoSFast top-1 & $\Delta$ top-1 & AdamW s & FANoSFast s & $\Delta$ time \\
\midrule
MNIST & 96.19 $\pm$ 0.37 & 97.08 $\pm$ 0.49 & 0.89 pp & 2.83 & 4.23 & 49.8\% \\
Fashion-MNIST & 85.98 $\pm$ 1.50 & 88.17 $\pm$ 0.67 & 2.20 pp & 2.66 & 4.30 & 61.6\% \\
CIFAR-10 & 51.17 $\pm$ 0.93 & 53.84 $\pm$ 1.38 & 2.67 pp & 3.53 & 5.53 & 56.8\% \\
\bottomrule
\end{tabular}

%% file: tables/gradnorm_ablation.tex
\begin{tabular}{llrrrr}
\toprule
Dataset & Interval & FANoSFast top-1 & $\Delta$ top-1 & FANoSFast s & $\Delta$ time \\
\midrule
MNIST & 1 & 97.24 $\pm$ 0.37 & 1.04 pp & 5.62 & 89.5\% \\
MNIST & 2 & 96.85 $\pm$ 0.40 & 0.65 pp & 4.45 & 58.3\% \\
MNIST & 4 & 96.41 $\pm$ 1.22 & 0.21 pp & 3.86 & 42.3\% \\
MNIST & 8 & 96.66 $\pm$ 0.54 & 0.47 pp & 4.63 & 72.7\% \\
Fashion-MNIST & 2 & 87.20 $\pm$ 1.81 & 1.22 pp & 4.59 & 60.0\% \\
Fashion-MNIST & 4 & 87.24 $\pm$ 1.16 & 1.26 pp & 4.25 & 48.4\% \\
CIFAR-10 & 2 & 51.35 $\pm$ 2.01 & 0.18 pp & 5.96 & 65.3\% \\
CIFAR-10 & 4 & 51.10 $\pm$ 1.87 & -0.07 pp & 5.72 & 60.3\% \\
\bottomrule
\end{tabular}

%% file: tables/pinn_preliminary.tex
\begin{tabular}{lrrr}
\toprule
Optimizer & residual metric & loss & seconds \\
\midrule
fanosv2 & 6.472e-09 & 5.483e-04 & 9.69 \\
adamw & 1.863e-08 & 1.193e-03 & 3.11 \\
sgd & 5.142e-08 & 2.883e-03 & 2.58 \\
rmsprop & 2.278e-04 & 5.010e-03 & 3.07 \\
\bottomrule
\end{tabular}

%% file: tables/eeg_preliminary.tex
\begin{tabular}{lrrr}
\toprule
Method & top-1 & $\Delta$ top-1 & $\Delta$ time \\
\midrule
FANoS auto & 45.33 $\pm$ 5.06 & 0.67 pp & 109.6\% \\
AdamW & 44.67 $\pm$ 4.47 & 0.00 pp & 0.0\% \\
\bottomrule
\end{tabular}

%% file: tables/scientific_preliminary.tex
\begin{tabular}{llrr}
\toprule
Task & Optimizer & metric & seconds \\
\midrule
rosenbrock100 & fanosv2 & 5.611e+00 & 2.05 \\
rosenbrock100 & adamw & 9.524e+00 & 0.52 \\
rosenbrock100 & rmsprop & 7.583e-01 & 0.51 \\
noisy\_small\_regression & fanosv2 & 5.714e-02 & 5.62 \\
noisy\_small\_regression & adamw & 5.318e-02 & 1.20 \\
noisy\_small\_regression & rmsprop & 6.056e-02 & 1.17 \\
ode\_exp\_fit & fanosv2 & 7.120e-03 & 1.85 \\
ode\_exp\_fit & adamw & 7.120e-03 & 0.40 \\
ode\_exp\_fit & rmsprop & 9.848e-03 & 0.39 \\
poisson\_pinn\_1d & fanosv2 & 1.445e-03 & 9.99 \\
poisson\_pinn\_1d & adamw & 1.863e-08 & 3.19 \\
poisson\_pinn\_1d & rmsprop & 2.278e-04 & 3.17 \\
sequence\_memory & fanosv2 & 1.000e+00 & 20.66 \\
sequence\_memory & adamw & 1.000e+00 & 11.99 \\
sequence\_memory & rmsprop & 7.266e-01 & 12.07 \\
\bottomrule
\end{tabular}

%% file: references.bib
@article{polyak1964some,
  author = {Polyak, Boris T.},
  title = {Some Methods of Speeding Up the Convergence of Iteration Methods},
  journal = {USSR Computational Mathematics and Mathematical Physics},
  volume = {4},
  number = {5},
  pages = {1--17},
  year = {1964}
}

@article{duchi2011adaptive,
  author = {Duchi, John and Hazan, Elad and Singer, Yoram},
  title = {Adaptive Subgradient Methods for Online Learning and Stochastic Optimization},
  journal = {Journal of Machine Learning Research},
  volume = {12},
  pages = {2121--2159},
  year = {2011}
}

@misc{tieleman2012lecture,
  author = {Tieleman, Tijmen and Hinton, Geoffrey},
  title = {Lecture 6.5---RMSProp: Divide the Gradient by a Running Average of Its Recent Magnitude},
  howpublished = {COURSERA: Neural Networks for Machine Learning},
  year = {2012}
}

@misc{kingma2014adam,
  author = {Kingma, Diederik P. and Ba, Jimmy},
  title = {Adam: A Method for Stochastic Optimization},
  year = {2014},
  eprint = {1412.6980},
  archivePrefix = {arXiv},
  primaryClass = {cs.LG}
}

@inproceedings{loshchilov2019decoupled,
  author = {Loshchilov, Ilya and Hutter, Frank},
  title = {Decoupled Weight Decay Regularization},
  booktitle = {International Conference on Learning Representations},
  year = {2019}
}

@inproceedings{shazeer2018adafactor,
  author = {Shazeer, Noam and Stern, Mitchell},
  title = {Adafactor: Adaptive Learning Rates with Sublinear Memory Cost},
  booktitle = {International Conference on Machine Learning},
  year = {2018}
}

@inproceedings{gupta2018shampoo,
  author = {Gupta, Vineet and Koren, Tomer and Singer, Yoram},
  title = {Shampoo: Preconditioned Stochastic Tensor Optimization},
  booktitle = {International Conference on Machine Learning},
  year = {2018}
}

@article{nose1984unified,
  author = {Nos{\'e}, Shuichi},
  title = {A Unified Formulation of the Constant Temperature Molecular Dynamics Methods},
  journal = {Journal of Chemical Physics},
  volume = {81},
  pages = {511--519},
  year = {1984}
}

@article{hoover1985canonical,
  author = {Hoover, William G.},
  title = {Canonical Dynamics: Equilibrium Phase-Space Distributions},
  journal = {Physical Review A},
  volume = {31},
  pages = {1695--1697},
  year = {1985}
}

@article{raissi2019physics,
  author = {Raissi, Maziar and Perdikaris, Paris and Karniadakis, George Em},
  title = {Physics-Informed Neural Networks: A Deep Learning Framework for Solving Forward and Inverse Problems Involving Nonlinear Partial Differential Equations},
  journal = {Journal of Computational Physics},
  volume = {378},
  pages = {686--707},
  year = {2019}
}

@article{lecun1998gradient,
  author = {LeCun, Yann and Bottou, L{\'e}on and Bengio, Yoshua and Haffner, Patrick},
  title = {Gradient-Based Learning Applied to Document Recognition},
  journal = {Proceedings of the IEEE},
  volume = {86},
  number = {11},
  pages = {2278--2324},
  year = {1998}
}

@misc{xiao2017fashion,
  author = {Xiao, Han and Rasul, Kashif and Vollgraf, Roland},
  title = {Fashion-MNIST: A Novel Image Dataset for Benchmarking Machine Learning Algorithms},
  year = {2017},
  eprint = {1708.07747},
  archivePrefix = {arXiv},
  primaryClass = {cs.LG}
}

@techreport{krizhevsky2009learning,
  author = {Krizhevsky, Alex},
  title = {Learning Multiple Layers of Features from Tiny Images},
  institution = {University of Toronto},
  year = {2009}
}

@article{goldberger2000physiobank,
  author = {Goldberger, Ary L. and Amaral, Luis A. N. and Glass, Leon and Hausdorff, Jeffrey M. and Ivanov, Plamen Ch. and Mark, Roger G. and Mietus, Joseph E. and Moody, George B. and Peng, Chung-Kang and Stanley, H. Eugene},
  title = {PhysioBank, PhysioToolkit, and PhysioNet: Components of a New Research Resource for Complex Physiologic Signals},
  journal = {Circulation},
  volume = {101},
  number = {23},
  pages = {e215--e220},
  year = {2000}
}

@article{schalk2004bci2000,
  author = {Schalk, Gerwin and McFarland, Dennis J. and Hinterberger, Thilo and Birbaumer, Niels and Wolpaw, Jonathan R.},
  title = {BCI2000: A General-Purpose Brain-Computer Interface System},
  journal = {IEEE Transactions on Biomedical Engineering},
  volume = {51},
  number = {6},
  pages = {1034--1043},
  year = {2004}
}
